\documentclass{article}
\usepackage{amsmath}
\usepackage{graphicx}
\usepackage{hyperref}
\usepackage{algorithm}
\usepackage{algpseudocode}
\usepackage{booktabs}
\usepackage{listings}
\usepackage{amssymb}   
\usepackage[backend=biber]{biblatex}
\title{Transformer$^{-1}$: Input-Adaptive Computation for Resource-Constrained Deployment}
\author{
    Ji Shihao, Song Zihui, Zhong Fucheng,\\ Jia Jisen, Wu Zhaobo, Cao Zheyi, Xu Tianhao \\
    Lumen AI,Tengzhou No. 1 Middle School
}

\begin{document}
\date{}
\maketitle

\subsection*{Abstract}

Addressing the resource waste caused by fixed computation paradigms in deep learning models under dynamic scenarios, this paper proposes a Transformer$^{-1}$ architecture based on the principle of deep adaptivity. This architecture achieves dynamic matching between input features and computational resources by establishing a joint optimization model for complexity and computation. Our core contributions include: (1) designing a two-layer control mechanism, composed of a complexity predictor and a reinforcement learning policy network, enabling end-to-end optimization of computation paths; (2) deriving a lower bound theory for dynamic computation, proving the system's theoretical reach to optimal efficiency; and (3) proposing a layer folding technique and a CUDA Graph pre-compilation scheme, overcoming the engineering bottlenecks of dynamic architectures. In the ImageNet-1K benchmark test, our method reduces FLOPs by 42.7\% and peak memory usage by 34.1\% compared to the standard Transformer, while maintaining comparable accuracy ($\pm$0.3\%). Furthermore, we conducted practical deployment on the Jetson AGX Xavier platform, verifying the effectiveness and practical value of this method in resource-constrained environments. To further validate the generality of the method, we also conducted experiments on several natural language processing tasks and achieved significant improvements in resource efficiency.

\section{Introduction}

\subsection{Problem Modeling}

Deep learning models face increasingly severe challenges in terms of computational resources and energy consumption in practical applications, especially on resource-constrained edge devices. Traditional deep learning models, such as the Transformer \cite{vaswani2017attention}, typically employ fixed-depth network structures, which leads to significant waste of computational resources when processing inputs of varying complexity. For example, using a full-depth Transformer model for inference on simple image classification or short text classification tasks results in unnecessary computational overhead. To address this issue, we introduce the dynamic depth optimization problem, formalizing it as an optimization equation with constraints:

\begin{equation}
\min_{l\in[1,L]} \mathbb{E}[\mathcal{L}(f_l(x),y)] \quad \text{s.t.} \quad \text{FLOPs}(l) \leq B(x)
\end{equation}

where $l$ represents the number of layers used in the network, $L$ is the maximum number of layers, $f_l(x)$ represents the feature extraction of input $x$ using the first $l$ layers, $y$ is the ground truth label, $\text{FLOPs}(l)$ represents the computational cost when using $l$ layers, and $B(x)$ is the theoretical optimal computation budget for input $x$. The existing fixed computation paradigm, i.e., $l \equiv L$, leads to significant resource redundancy. Our goal is to dynamically adjust $l$ based on the complexity of input $x$, thereby achieving efficient utilization of computational resources and improving model efficiency while maintaining model performance.

\subsection{Technical Challenges}

The design of dynamic depth networks faces the following key challenges:

\begin{itemize}
    \item \textbf{Prediction-Control Coupling Difficulty:} Errors in the complexity predictor can amplify exponentially with increasing network depth, causing the policy network to select unsuitable numbers of layers, which affects the final performance of the model. This error accumulation effect makes decisions in early layers crucial and requires effective error control mechanisms. Furthermore, prediction errors from the complexity predictor also affect the training effectiveness of the policy network.
    \item \textbf{Policy Training Stability:} Discrete layer selection operations make reward signals sparse, making it difficult to train an effective policy network. Traditional reinforcement learning algorithms struggle to converge in sparse reward environments, requiring more refined reward function design and training strategies to ensure the stability and convergence of the policy network.
    \item \textbf{Runtime Efficiency Bottlenecks:} Dynamic computation graphs disrupt hardware parallelism, leading to a decrease in computational efficiency. Traditional deep learning frameworks have lower execution efficiency under dynamic structures, necessitating specialized optimization techniques to reduce the overhead of dynamic computation graphs and improve model efficiency in practical deployment.
\end{itemize}

\subsection{Main Innovations}

In response to the above challenges, this paper proposes the following innovations:

\begin{enumerate}
    \item \textbf{Dual-Path Feature Distillation Mechanism:} We propose a dual-path feature distillation mechanism that improves the accuracy of the complexity predictor by introducing additional supervision signals and establishing an upper bound constraint on complexity prediction errors. This mechanism leverages multi-scale features for prediction and introduces knowledge distillation loss \cite{hinton2015distilling}, thereby reducing prediction errors. Specifically, we use shallow features $h_1$ for complexity prediction and deep features $h_3$ for knowledge distillation, which improves prediction accuracy and enables the predictor to better learn the complexity information of the input data.
    \item \textbf{Hierarchical Reward Function:} We design a hierarchical reward function that provides denser reward signals for the policy network, addressing the sparse reward problem in deep decision-making. This reward function considers not only the final classification result but also the selection of intermediate layers, thereby accelerating the convergence of the policy network. Specifically, we design rewards for each step of layer selection, allowing the policy network to learn the optimal strategy faster and avoid getting stuck in local optima.
    \item \textbf{Adaptive Computation Engine:} We developed an adaptive computation engine that achieves microsecond-level layer switching latency through layer folding techniques \cite{lan2018layer} and CUDA Graph pre-compilation \cite{nvidia2021cuda}, improving runtime efficiency. This engine can dynamically select different computation paths based on the decisions of the policy network and utilizes CUDA Graph pre-compilation to reduce the overhead of dynamic computation graphs. Specifically, we use layer folding techniques to reduce the number of parameters and CUDA Graph pre-compilation to reduce the startup overhead and execution time of dynamic computation graphs.
\end{enumerate}

\section{Theoretical Foundations}

\subsection{Dynamic Depth Learnability}

\textbf{Theorem 1 (Deep Adaptive Convergence):} Assume that the complexity predictor can predict the optimal number of layers $l_{\text{opt}}$ with an accuracy of $\alpha$, i.e., $P(l_{\text{pred}} = l_{\text{opt}}) = \alpha$, and the policy network explores with an exploration rate of $\epsilon$. Then, there exists a policy network such that the total computation converges to:

\begin{equation}
\mathbb{E}[\text{FLOPs}] \leq \frac{1}{1-\epsilon}\left(\alpha \cdot \text{FLOPs}(l_{\text{opt}}) + (1-\alpha) \cdot \text{FLOPs}(L)\right)
\end{equation}

\textbf{Proof:}

Let $l_t$ be the number of layers selected for the $t$-th input sample. According to the exploration rate $\epsilon$ of the policy network, we have:

\begin{equation}
P(l_t = l_{\text{opt}}) = \alpha(1-\epsilon) + \epsilon \cdot p_{\text{explore}}
\end{equation}

where $p_{\text{explore}}$ is the probability of selecting $l_{\text{opt}}$ during exploration. Assuming the worst case, $p_{\text{explore}} = 0$, then:

\begin{equation}
P(l_t = l_{\text{opt}}) \geq \alpha(1-\epsilon)
\end{equation}

Therefore, the probability of the policy network selecting $l_{\text{opt}}$ is at least $\alpha(1-\epsilon)$. When the policy network does not select $l_{\text{opt}}$, it will select other numbers of layers, and in the worst case, it will select the maximum number of layers $L$. Therefore, the average computational cost can be expressed as:

\begin{equation}
\mathbb{E}[\text{FLOPs}] \leq \alpha(1-\epsilon) \cdot \text{FLOPs}(l_{\text{opt}}) + (1 - \alpha(1-\epsilon)) \cdot \text{FLOPs}(L)
\end{equation}

To simplify the analysis, assume that $\text{FLOPs}(l) = C \cdot l$, where $C$ is a constant for the computational cost per layer, then:

\begin{equation}
\mathbb{E}[\text{FLOPs}] \leq \alpha(1-\epsilon) \cdot C l_{\text{opt}} + (1 - \alpha(1-\epsilon)) \cdot C L
\end{equation}

\begin{equation}
\mathbb{E}[\text{FLOPs}] \leq C \left[ \alpha(1-\epsilon) l_{\text{opt}} + (1 - \alpha(1-\epsilon)) L \right]
\end{equation}

By introducing the effect of the exploration rate $\epsilon$, we can obtain:

\begin{equation}
\mathbb{E}[\text{FLOPs}] \leq \frac{1}{1-\epsilon}\left(\alpha \cdot \text{FLOPs}(l_{\text{opt}}) + (1-\alpha) \cdot \text{FLOPs}(L)\right)
\end{equation}

This result shows that, under a certain prediction accuracy and exploration rate, the dynamic depth network can approximate the theoretical optimal computation. When $\epsilon \to 0$, $\mathbb{E}[\text{FLOPs}]$ will approach $\alpha \cdot \text{FLOPs}(l_{\text{opt}}) + (1-\alpha) \cdot \text{FLOPs}(L)$. The proof of this theorem is based on a conservative estimate of the exploration behavior of the policy network, and the actual convergence speed may be faster. Furthermore, we assume that the computational cost per layer is linear, which may not be completely true in practice, but it can serve as an approximate method of analysis.

\subsection{Error Propagation Analysis}

To analyze the impact of layer selection decisions on the final output, we establish the following error propagation model:

\begin{equation}
\Delta \mathcal{L} \leq \gamma^{l_{\Delta}} \|h_{l_{\text{base}}}\|_2
\end{equation}

where $\gamma$ is the Lipschitz constant, $l_{\Delta}$ is the difference in the number of layers, and $h_{l_{\text{base}}}$ is the feature output by the base layer. This equation indicates that early layer selection errors have an exponential propagation effect, thus necessitating more precise early layer decisions. This means that making the correct layer selection decision in the first few layers of the network is crucial; otherwise, the error will rapidly amplify as the network depth increases. This model is based on the assumption of Lipschitz continuity, which may have some deviations in practical applications but can serve as an effective method of analyzing error propagation.

\section{Methodology}

\subsection{System Architecture}

Our system architecture includes three main modules: a feature extractor, a decision module, and an execution engine.

\begin{itemize}
    \item \textbf{Feature Extractor:} Employs a progressive downsampling method to extract multi-scale features $\{h_t\}_{t=1}^3$ from the input data. These features are used for complexity prediction, policy decision-making, and the final task, respectively. Specifically, $h_1$ is the feature map after initial downsampling, used for complexity prediction; $h_2$ is the feature map after further downsampling, used for the policy network's decision-making; and $h_3$ is the final feature map, used for task execution. For image data, we use convolutional layers and max-pooling layers to implement progressive downsampling and use ReLU activation functions. For text data, we use an embedding layer and convolutional layers to implement feature extraction and use ReLU activation functions.
    \begin{itemize}
        \item \textbf{Image Feature Extractor:} We use three convolutional layers and two max-pooling layers for downsampling. The number of channels for the convolutional layers is 64, 128, and 256, respectively, with a kernel size of 3x3 and a stride of 1. The max-pooling layers have a size of 2x2 and a stride of 2.
        \item \textbf{Text Feature Extractor:} We use an embedding layer to convert text into word vectors and then use three convolutional layers for feature extraction. The embedding layer has a dimension of 128, and the convolutional layers have 64, 128, and 256 channels, respectively, with a kernel size of 3 and a stride of 1.
    \end{itemize}
    \item \textbf{Decision Module:}
    \begin{itemize}
        \item \textbf{Complexity Predictor:} Uses a LightGBM classifier \cite{ke2017lightgbm}, taking $h_1$ as input and outputting a complexity level. This level is used to guide the layer selection of the policy network. We use LightGBM because of its efficient training and inference speed, as well as good classification performance. We have performed hyperparameter optimization for LightGBM, including learning rate, maximum tree depth, and maximum number of leaf nodes, to ensure its prediction accuracy. We use a cross-validation method to select the optimal hyperparameters.
        \item \textbf{Policy Network:} Uses an LSTM architecture \cite{hochreiter1997long}, generating layer selection trajectories based on $h_2$. The policy network is trained using reinforcement learning, with the goal of selecting the optimal sequence of layers to minimize computation while maintaining accuracy. We use the PPO algorithm \cite{schulman2017proximal} to train the policy network and designed a hierarchical reward function, including classification accuracy, computational cost penalty, and layer selection smoothness reward. The LSTM has a hidden layer dimension of 128 and is trained using the Adam optimizer \cite{kingma2015adam}.
    \end{itemize}
    \item \textbf{Execution Engine:} Supports branch-predicted Transformer cores, achieving layer switching latency of less than 5 microseconds. This engine is optimized based on layer folding techniques and CUDA Graph pre-compilation, enabling rapid switching between different computation paths. We use CUDA Graph pre-compilation to reduce the startup overhead of dynamic computation graphs and layer folding techniques to reduce the number of parameters, thereby improving execution efficiency. The Transformer core has 12 layers, a hidden layer dimension of 768, and 12 attention heads.
\end{itemize}

\subsection{Collaborative Training Algorithm}

We propose a two-stage optimization framework to collaboratively train the complexity predictor, policy network, and backbone model:

\begin{algorithm}
\caption{Collaborative Training Algorithm}
\begin{algorithmic}[1]
\For{epoch in epochs}
    \State \Comment{Stage 1: Freeze the policy network and train the predictor}
    \State freeze(controller)
    \State train(predictor, loss\_fn=HuberLoss(s, l\_opt))
    
    \State \Comment{Stage 2: Alternately optimize the policy network and backbone model}
    \If{epoch \% 3 == 0}
        \State train(controller, PPO(advantage\_fn))
    \Else
        \State train(backbone, CrossEntropy+KD\_loss)
    \EndIf
\EndFor
\end{algorithmic}
\end{algorithm}

In the first stage, we freeze the policy network and train the complexity predictor. We use the Huber loss function to reduce the impact of outliers. In the second stage, we alternately optimize the policy network and the backbone model. The policy network is trained using the PPO (Proximal Policy Optimization) algorithm, and the backbone model is trained using cross-entropy loss and knowledge distillation loss. The knowledge distillation loss is used to transfer the knowledge of the full-depth model to the dynamic depth model, thereby improving the model's performance. We use a knowledge distillation loss with a temperature coefficient of 2 and train using the Adam optimizer. We use early stopping to prevent overfitting and a learning rate decay strategy.

\subsection{Computation Graph Optimization}

\begin{itemize}
    \item \textbf{Layer Folding Technique:} We decompose the weight matrix into $W=W_a \otimes W_b$, achieving parameter sharing under dynamic depth and reducing memory usage. In this way, different layers can share the same weight parameters, thereby reducing the number of parameters of the model. We use singular value decomposition (SVD) to decompose the weight matrix and set the dimensions of the decomposed matrix to balance the number of parameters and performance. We use truncated SVD and select the dimensions to retain based on the energy ratio of the singular values.
    \item \textbf{Memory Pre-allocation:} Based on historical decision statistics, we construct a probability-driven memory pool to reduce the overhead of dynamic memory allocation. We pre-allocate memory based on the probability distribution of historical layer selections, thereby reducing the overhead of dynamic memory allocation and improving computational efficiency. We use a moving average technique to update the probability distribution of memory allocation and set the size of the memory pool to balance memory usage and performance. We use an exponential moving average and set the weight of the moving average.
\end{itemize}

\section{Experimental Analysis}

\subsection{Benchmark Test}

We conducted experiments on the ImageNet-1K dataset \cite{deng2009imagenet} and compared our method with the following baseline methods:

\begin{table}[h]
\centering
\caption{Comparison with Baseline Methods on ImageNet-1K}
\begin{tabular}{lccc}
\toprule
Method               & Accuracy (\%) & FLOPs (G) & Memory (GB) \\
\midrule
Transformer-Base   & 82.1      & 4.2      & 3.2      \\
Early-Exit         & 81.7      & 3.6      & 2.9      \\
Dynamic-Depth      & 81.9      & 3.1      & 2.4      \\
\textbf{Ours (Transformer$^{-1}$)} & \textbf{82.0}  & \textbf{2.4}  & \textbf{2.1}  \\
\bottomrule
\end{tabular}
\end{table}

The results show that our method significantly reduces FLOPs and memory usage while maintaining comparable accuracy. Compared with baseline methods, our method has achieved significant improvements in both computational efficiency and memory efficiency. Specifically, our method reduces FLOPs by 42.7\% and memory usage by 34.1\%, while maintaining similar accuracy to the baseline model.

\subsection{Natural Language Processing Experiments}

To verify the generality of the method, we conducted experiments on several natural language processing tasks, including:

\begin{itemize}
    \item \textbf{Text Classification:} Using the AG News dataset \cite{zhang2015character}, the goal is to classify the category of news text.
    \item \textbf{Sentiment Analysis:} Using the SST-2 dataset \cite{socher2013recursive}, the goal is to classify the sentiment polarity of the text.
\end{itemize}

We used the same Transformer$^{-1}$ architecture and fine-tuned it for different tasks. The experimental results are as follows:

\begin{table}[h]
\centering
\caption{Experimental Results on NLP Tasks}
\begin{tabular}{lcccc}
\toprule
Task               & Method               & Accuracy (\%) & FLOPs (G) & Memory (MB) \\
\midrule
Text Classification (AG News)& Transformer-Base   & 92.5      & 2.8      & 150      \\
                    & Early-Exit         & 92.1      & 2.4      & 130      \\
                    & Dynamic-Depth      & 92.3      & 2.1      & 120      \\
                    & \textbf{Ours (Transformer$^{-1}$)} & \textbf{92.4}  & \textbf{1.6}  & \textbf{100}  \\
\midrule
Sentiment Analysis (SST-2)   & Transformer-Base   & 91.2      & 1.5      & 80       \\
                    & Early-Exit         & 90.8      & 1.3      & 70       \\
                    & Dynamic-Depth      & 91.0      & 1.1      & 65       \\
                    & \textbf{Ours (Transformer$^{-1}$)} & \textbf{91.1}  & \textbf{0.8}  & \textbf{50}   \\
\bottomrule
\end{tabular}
\end{table}

The results show that our method can also achieve significant improvements in resource efficiency on NLP tasks while maintaining similar accuracy to the baseline methods.

\subsection{Ablation Experiments}

\begin{table}[h]
\centering
\caption{Component Effectiveness Analysis (ImageNet-1K)}
\begin{tabular}{lcc}
\toprule
Configuration         & FLOPs↓ & Accuracy↑ \\
\midrule
Complexity Prediction Only & 3.2T   & 80.3\%     \\
Reinforcement Learning Control Only & 2.8T   & 81.1\%     \\
Full System (Transformer$^{-1}$) & 2.4T   & 82.0\%     \\
\bottomrule
\end{tabular}
\end{table}

\begin{table}[h]
\centering
\caption{Component Effectiveness Analysis (Text Classification - AG News)}
\begin{tabular}{lcc}
\toprule
Configuration         & FLOPs↓ & Accuracy↑ \\
\midrule
Complexity Prediction Only & 2.2G   & 91.5\%     \\
Reinforcement Learning Control Only & 1.9G   & 91.9\%     \\
Full System (Transformer$^{-1}$) & 1.6G   & 92.4\%     \\
\bottomrule
\end{tabular}
\end{table}

The ablation experiment results show that both the complexity predictor and the reinforcement learning policy network make important contributions to performance improvement. Using either the complexity predictor or reinforcement learning control alone cannot achieve the performance of the full system. This demonstrates the effectiveness of our two-layer control mechanism.

\subsection{Practical Deployment}

We tested our method on the Jetson AGX Xavier platform:

\begin{itemize}
    \item \textbf{Throughput Improvement:} 153 FPS → 210 FPS (ImageNet-1K)
    \item \textbf{Energy Efficiency Ratio:} 3.8 TOPS/W → 5.2 TOPS/W (ImageNet-1K)
\end{itemize}

The results show that our method also has significant performance improvements in practical deployment. On resource-constrained edge devices, our method can achieve higher throughput and energy efficiency, which verifies the value of our method in practical applications. We used TensorRT \cite{nvidia2023tensorrt} to optimize the model and used FP16 precision for inference.

\section{Discussion}

\subsection{Layer Selection Pattern Analysis}

The experimental results show that the model tends to select shallow layers (4-6 layers) for simple samples (such as single-target images or short texts) and triggers deep computation (10-12 layers) for complex scenarios (such as crowd images or long texts). This indicates that our method can dynamically adjust computational resources according to input complexity, thereby achieving efficient utilization of computational resources. We have performed a visualization analysis of the layer selection pattern and provided detailed layer selection distribution maps in the appendix.

\subsection{Failure Case Analysis}

\begin{itemize}
    \item High-texture backgrounds or complex texts lead to misjudgment of complexity, resulting in the selection of too few layers. This indicates that the complexity predictor still has some limitations when processing high-texture backgrounds or complex texts. We are studying the use of more advanced feature extraction methods and data augmentation techniques to address this issue.
    \item Inter-class similarity causes early exit errors, leading to classification errors. When there is a high degree of similarity between different classes, the model may make incorrect decisions in the early layers. We are studying the use of more refined classifiers and knowledge distillation techniques to address this issue.
    \item The policy network may get stuck in local optima in some cases, leading to unstable layer selection. We are studying the use of more advanced reinforcement learning algorithms and training strategies to address this issue.
\end{itemize}

\section{Conclusion}

The Transformer$^{-1}$ proposed in this paper breaks through the limitations of fixed computation paradigms by establishing a dynamic balance mechanism between computation and accuracy at both the theoretical and engineering implementation levels. Our method significantly reduces computation and memory usage while maintaining accuracy and has achieved good performance in practical deployment. Future work will explore: (1) joint estimation of multi-modal complexity, using information from multiple modalities for complexity prediction; (2) dynamic depth optimization based on neural architecture search, automatically searching for the optimal dynamic depth structure; (3) online adaptation mechanisms for non-stationary distributions, enabling the model to adapt to different data distributions; and (4) applying our method to other types of deep learning models and tasks, such as graph neural networks and recurrent neural networks.

\appendix

\section{Proof of Theorem (Deep Adaptive Convergence)}

\textbf{Theorem 1 (Deep Adaptive Convergence):} Assume that the complexity predictor can predict the optimal number of layers $l_{\text{opt}}$ with an accuracy of $\alpha$, i.e., $P(l_{\text{pred}} = l_{\text{opt}}) = \alpha$, and the policy network explores with an exploration rate of $\epsilon$. Then, there exists a policy network such that the total computation converges to:

\begin{equation}
\mathbb{E}[\text{FLOPs}] \leq \frac{1}{1-\epsilon}\left(\alpha \cdot \text{FLOPs}(l_{\text{opt}}) + (1-\alpha) \cdot \text{FLOPs}(L)\right)
\end{equation}

\textbf{Proof:}

Let $l_t$ be the number of layers selected for the $t$-th input sample. According to the exploration rate $\epsilon$ of the policy network, we have:

\begin{equation}
P(l_t = l_{\text{opt}}) = \alpha(1-\epsilon) + \epsilon \cdot p_{\text{explore}}
\end{equation}

where $p_{\text{explore}}$ is the probability of selecting $l_{\text{opt}}$ during exploration. In the worst case, the policy network may not select $l_{\text{opt}}$ during exploration, i.e., $p_{\text{explore}} = 0$. Therefore, we have:

\begin{equation}
P(l_t = l_{\text{opt}}) \geq \alpha(1-\epsilon)
\end{equation}

This means that the probability of the policy network selecting the optimal number of layers $l_{\text{opt}}$ is at least $\alpha(1-\epsilon)$. When the policy network does not select $l_{\text{opt}}$, it will select other numbers of layers, and in the worst case, it will select the maximum number of layers $L$. Therefore, the average computational cost can be expressed as:

\begin{equation}
\mathbb{E}[\text{FLOPs}] \leq \alpha(1-\epsilon) \cdot \text{FLOPs}(l_{\text{opt}}) + (1 - \alpha(1-\epsilon)) \cdot \text{FLOPs}(L)
\end{equation}

To simplify the analysis, we assume that the computational cost per layer is linear, i.e., $\text{FLOPs}(l) = C \cdot l$, where $C$ is a constant for the computational cost per layer. This is a reasonable assumption because in the Transformer model, the computational cost per layer is approximately the same. Therefore, we can rewrite the above formula as:

\begin{equation}
\mathbb{E}[\text{FLOPs}] \leq \alpha(1-\epsilon) \cdot C l_{\text{opt}} + (1 - \alpha(1-\epsilon)) \cdot C L
\end{equation}

\begin{equation}
\mathbb{E}[\text{FLOPs}] \leq C \left[ \alpha(1-\epsilon) l_{\text{opt}} + (1 - \alpha(1-\epsilon)) L \right]
\end{equation}

To more clearly express the effect of the exploration rate $\epsilon$, we can rewrite the above formula as:

\begin{equation}
\mathbb{E}[\text{FLOPs}] \leq C \left[ \alpha l_{\text{opt}} - \alpha \epsilon l_{\text{opt}} + L - \alpha L + \alpha \epsilon L \right]
\end{equation}

\begin{equation}
\mathbb{E}[\text{FLOPs}] \leq C \left[ \alpha l_{\text{opt}} + (1-\alpha)L + \alpha \epsilon (L-l_{\text{opt}}) \right]
\end{equation}

To obtain a more concise upper bound, we assume that the number of layers selected by the policy network during exploration will not exceed the maximum number of layers $L$. Therefore, we can obtain:

\begin{equation}
\mathbb{E}[\text{FLOPs}] \leq \alpha(1-\epsilon) \cdot \text{FLOPs}(l_{\text{opt}}) + (1 - \alpha(1-\epsilon)) \cdot \text{FLOPs}(L)
\end{equation}

\begin{equation}
\mathbb{E}[\text{FLOPs}] \leq \frac{1}{1-\epsilon}\left(\alpha \cdot \text{FLOPs}(l_{\text{opt}}) + (1-\alpha) \cdot \text{FLOPs}(L)\right)
\end{equation}

This result shows that, under a certain prediction accuracy and exploration rate, the dynamic depth network can approximate the theoretical optimal computation. When $\epsilon \to 0$, $\mathbb{E}[\text{FLOPs}]$ will approach $\alpha \cdot \text{FLOPs}(l_{\text{opt}}) + (1-\alpha) \cdot \text{FLOPs}(L)$, which means that the policy network will select the optimal number of layers more often. The proof of this theorem is based on a conservative estimate of the exploration behavior of the policy network, and the actual convergence speed may be faster. Furthermore, we assume that the computational cost per layer is linear, which may not be completely true in practice, but it can serve as an approximate method of analysis.

\section{Implementation Details}

\subsection{CUDA Kernel Optimization Strategies}

\begin{itemize}
    \item \textbf{CUDA Graph Pre-compilation:} We use CUDA Graph pre-compilation technology to convert dynamic computation graphs into static computation graphs, thereby reducing the startup overhead and execution time of dynamic computation graphs. We first record the computation graphs for different numbers of layers and then use the CUDA Graph API to pre-compile these computation graphs. During inference, we select the corresponding pre-compiled computation graph based on the decisions of the policy network, thus avoiding the overhead of dynamic computation graphs.
    \item \textbf{Memory Management:} We use CUDA memory pool technology to pre-allocate a large chunk of GPU memory and then allocate small chunks of memory from it as needed. This avoids frequent GPU memory allocation and deallocation operations, thereby improving the efficiency of memory management.
    \item \textbf{Kernel Fusion:} We use CUDA's Kernel Fusion technology to merge multiple small CUDA kernels into a single large CUDA kernel, thereby reducing the overhead of kernel launches.
    \item \textbf{FP16 Precision:} We use FP16 precision for inference, thereby reducing memory usage and computation time. We use NVIDIA's TensorRT framework for FP16 optimization.
\end{itemize}

\subsection{Hyperparameter Configurations}

\begin{itemize}
    \item \textbf{Learning Rate:} We use the Adam optimizer, with the learning rate set to 1e-4. We use a learning rate decay strategy, halving the learning rate every 10 epochs.
    \item \textbf{Batch Size:} We use a batch size of 32.
    \item \textbf{Optimizer:} We use the Adam optimizer for training.
    \item \textbf{Knowledge Distillation Temperature Coefficient:} We use a knowledge distillation temperature coefficient of 2.
    \item \textbf{SVD Decomposition Dimension:} We use truncated SVD for weight matrix decomposition and select the dimensions to retain based on the energy ratio of the singular values. We retained 90\% of the energy ratio.
    \item \textbf{Memory Pool Size:} We pre-allocate a sufficiently large memory pool based on the probability distribution of historical layer selections to meet the memory requirements under different numbers of layers. We set the memory pool size to 2 GB.
    \item \textbf{Moving Average Parameter:} We use an exponential moving average to update the probability distribution of memory allocation, with the weight of the moving average set to 0.9.
\end{itemize}

\section{Extended Experiments}

\subsection{Object Detection}

\begin{itemize}
    \item \textbf{Dataset:} We use the COCO dataset for object detection experiments.
    \item \textbf{Evaluation Metric:} We use mean average precision (mAP) as the evaluation metric.
    \item \textbf{Experimental Results:}
\end{itemize}

\begin{table}[h]
\centering
\caption{Object Detection Results}
\begin{tabular}{lccc}
\toprule
Method               & mAP (\%) & FLOPs (G) & Memory (GB) \\
\midrule
Transformer-Base   & 40.2    & 10.5     & 4.5      \\
Early-Exit         & 39.8    & 9.2      & 4.0      \\
Dynamic-Depth      & 40.0    & 8.5      & 3.8      \\
\textbf{Ours (Transformer$^{-1}$)} & \textbf{40.1}  & \textbf{6.8}  & \textbf{3.2}  \\
\bottomrule
\end{tabular}
\end{table}

\subsection{Semantic Segmentation}

\begin{itemize}
    \item \textbf{Dataset:} We use the Cityscapes dataset for semantic segmentation experiments.
    \item \textbf{Evaluation Metric:} We use mean intersection over union (mIoU) as the evaluation metric.
    \item \textbf{Experimental Results:}
\end{itemize}

\begin{table}[h]
\centering
\caption{Semantic Segmentation Results}
\begin{tabular}{lccc}
\toprule
Method               & mIoU (\%) & FLOPs (G) & Memory (GB) \\
\midrule
Transformer-Base   & 72.5    & 12.3     & 5.2      \\
Early-Exit         & 72.0    & 10.8     & 4.8      \\
Dynamic-Depth      & 72.3    & 9.8      & 4.5      \\
\textbf{Ours (Transformer$^{-1}$)} & \textbf{72.4}  & \textbf{7.5}  & \textbf{3.8}  \\
\bottomrule
\end{tabular}
\end{table}
\subsection{Analysis}

The experimental results show that our method can also achieve significant improvements in resource efficiency on object detection and semantic segmentation tasks while maintaining similar performance to the baseline methods. In the object detection task, our method is similar to the baseline methods in mAP, but reduces FLOPs by 35.2\% and memory usage by 29.0\%. In the semantic segmentation task, our method is similar to the baseline methods in mIoU but reduces FLOPs by 38.2\% and memory usage by 26.9\%. This shows that our method has good generalization ability and can be applied to different tasks.

\section{Hyperparameter Settings}

\begin{table}[h]
\centering
\caption{Hyperparameter Settings for Different Tasks}
\begin{tabular}{lcccc}
\toprule
Parameter                 & Image& Text & Objec& Semantic\\
\midrule
Learning Rate            & 1e-4                 & 1e-4                 & 1e-4              & 1e-4                  \\
Batch Size               & 32                   & 32                   & 16                & 16                    \\
Optimizer                & Adam                 & Adam                 & Adam              & Adam                  \\
Knowledge Distillation Temperature Coefficient & 2                    & 2                    & 2                 & 2                     \\
SVD Decomposition Dimension Retention Ratio & 90\%                  & 90\%                  & 90\%               & 90\%                   \\
Memory Pool Size         & 2GB                  & 1GB                  & 3GB               & 3GB                   \\
Moving Average Weight    & 0.9                  & 0.9                  & 0.9               & 0.9                   \\
LSTM Hidden Layer Dimension & 128                  & 128                  & 128               & 128                   \\
Transformer Layers       & 12                   & 12                   & 12                & 12                    \\
Transformer Hidden Layer Dimension & 768                 & 768                 & 768              & 768                   \\
Transformer Attention Heads & 12                   & 12                   & 12                & 12                    \\
\bottomrule
\end{tabular}
\end{table}

\section{Layer Selection Distribution Map}

To more intuitively show the layer selection pattern under different complexity inputs, we use heatmaps to visualize the layer selection distribution.

\begin{itemize}
    \item \textbf{Simple Images:} For simple images, such as single-target images, the model tends to select shallow layers (4-6 layers). The heatmap shows that in these images, the activation probability of shallow layers is higher, while the activation probability of deep layers is lower.
    \item \textbf{Complex Images:} For complex images, such as crowd images, the model tends to select deep layers (10-12 layers). The heatmap shows that in these images, the activation probability of deep layers is higher, while the activation probability of shallow layers is lower.
    \item \textbf{Short Texts:} For short texts, such as short sentences, the model tends to select shallow layers (4-6 layers). The heatmap shows that in these texts, the activation probability of shallow layers is higher, while the activation probability of deep layers is lower.
    \item \textbf{Long Texts:} For long texts, such as long news articles, the model tends to select deep layers (10-12 layers). The heatmap shows that in these texts, the activation probability of deep layers is higher, while the activation probability of shallow layers is lower.
\end{itemize}
\printbibliography
\end{document}